\documentclass[10pt, conference, compsocconf]{IEEEtran}
\usepackage{cite}

\usepackage{graphicx}
\usepackage{dblfloatfix}
\usepackage{float}
\usepackage{cuted}
\DeclareGraphicsExtensions{.pdf,.jpg,.png}
\usepackage{amssymb}

\usepackage[cmex10]{amsmath}
\usepackage{color}

\newcommand{\mshc}[1]{}
\usepackage{algorithmic}
\usepackage{array}
\usepackage{booktabs}
\usepackage{mdwmath}
\usepackage{mdwtab}

\usepackage[caption=false,font=footnotesize]{subfig}

\usepackage{url}

\newlength{\bibitemsep}
\newlength{\bibparskip}
\let\oldthebibliography\thebibliography
\renewcommand\thebibliography[1]{%
	\oldthebibliography{#1}%
	\setlength{\parskip}{\bibitemsep}%
	\setlength{\itemsep}{\bibparskip}%
}
\begin{document}
\title{Tracking the Evolution of Words with Time-reflective Text Representations}


\author{\IEEEauthorblockN{Roberto Camacho Barranco\IEEEauthorrefmark{1}, Raimundo F. Dos Santos\IEEEauthorrefmark{2}, M. Shahriar Hossain\IEEEauthorrefmark{3}, Monika Akbar\IEEEauthorrefmark{4}}
\IEEEauthorblockA{Department of Computer Science\\
University of Texas at El Paso, El Paso, TX, USA\\
Email: \IEEEauthorrefmark{1}rcamachobarranco@utep.edu,
\IEEEauthorrefmark{3}mhossain@utep.edu,
\IEEEauthorrefmark{4}makbar@utep.edu}
\IEEEauthorblockA{U. S. Army Corps of Engineers, Alexandria, VA, USA\\
Email: \IEEEauthorrefmark{2}raimundo.f.dossantos@erdc.dren.mil}
}

\maketitle

\begin{abstract}
More than 80\% of today's data is unstructured in nature, and these unstructured datasets evolve over time. A large part of these datasets are text documents generated by media outlets, scholarly articles in digital libraries, findings from scientific and professional communities, and social media. Vector space models were developed to analyze text data using data mining and machine learning algorithms. While ample vector space models exist for text data, the evolutionary aspect of ever changing text corpora is still missing in vector-based representations. The advent of word embeddings has enabled us to create a contextual vector space, but the embeddings fail to consider the temporal aspects of the feature space successfully. This paper presents an approach to include temporal aspects in feature spaces. The inclusion of the time aspect in the feature space provides vectors for every natural language element, such as words or entities, at every timestamp. Such temporal word vectors allow us to track how the meaning of a word changes over time, by studying the changes in its neighborhood. Moreover, a time-reflective text representation will pave the way to a new set of text analytic abilities involving time series for text collections.

In this paper, we present a time-reflective vector space model for temporal text data that is able to capture short and long-term changes in the meaning of words. We compare our approach with the limited literature on dynamic embeddings. We present qualitative and quantitative evaluations using the tracking of semantic evolution as the target application.
\end{abstract}


% \fi
\IEEEpeerreviewmaketitle

% \thanks{This material is based upon work supported by the U.S. Army Engineering Research and Development Center under Contract No. W9132V-15-C-0006.}

\section{Introduction}\label{sec:intro}
\begin{figure}
	\centering
	\includegraphics[width=\columnwidth]{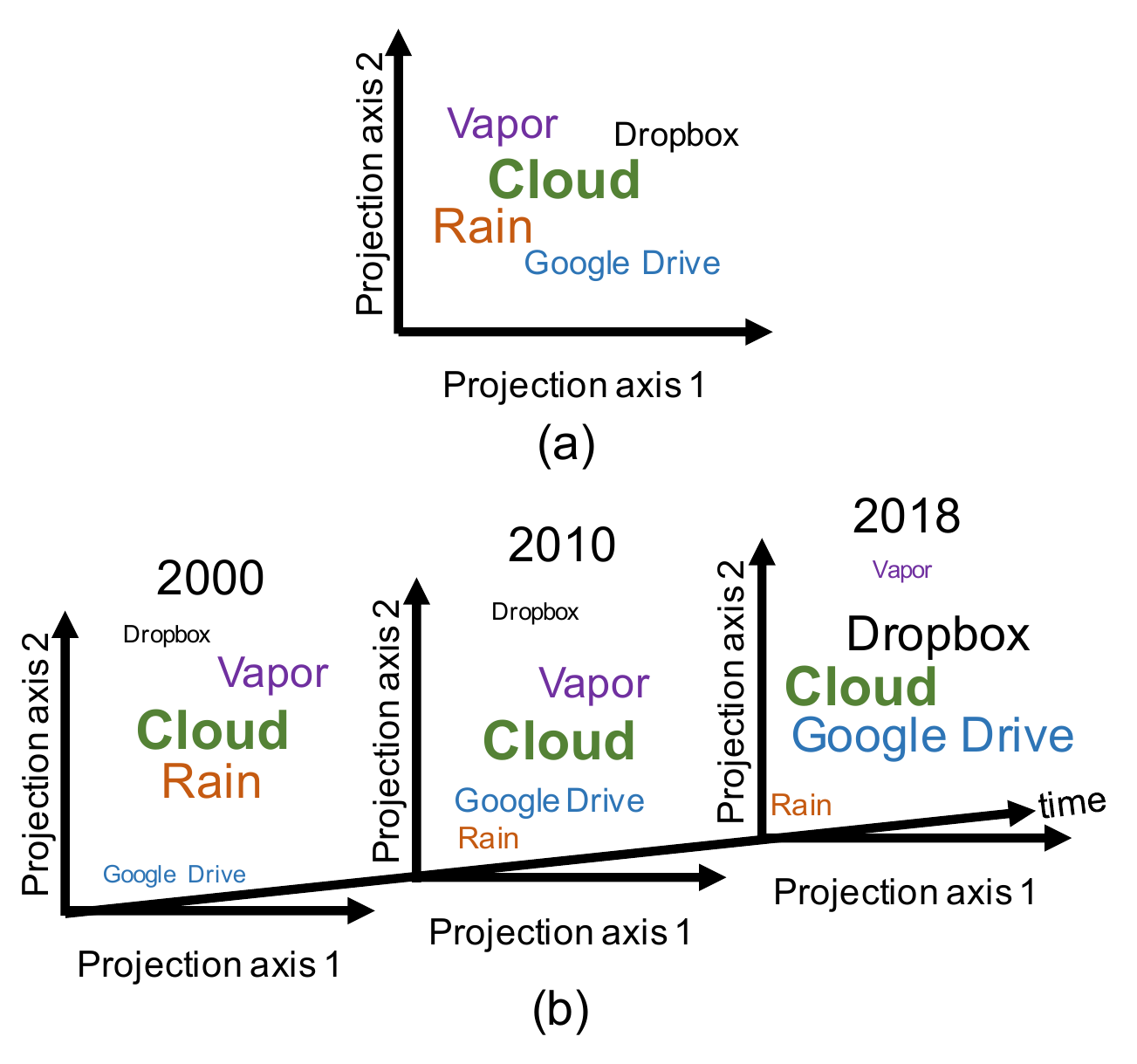}	
	\caption{Difference between static representations and time-reflective representations.}
	\label{fig:ipad}
\end{figure}
With the high volumes of text data generated by every sector of society, the necessity to consider text publications as an evolving stream of data is increasing. There is no doubt that the existing abilities to transform unstructured text collections into a structured representation provide us with ample analytic opportunities by leveraging many data mining and machine learning algorithms that have been designed exclusively for structured data. However, to serve an even larger set of analytic needs, modern text mining is slowly drifting toward analysis of temporal aspects of text \cite{hamilton_diachronic_2016, tang_state---art_2018}. The ability to represent unstructured text with a temporal context is in its infant stage. This paper discusses the needs and expected properties of a time-reflective representation of text data along with a preliminary implementation that reveals the potential of such time-reflective representations.

\begin{table*}	
	\caption{Feature comparison between vector space models}
	\label{table:comparison}
	\begin{tabular}{|c|c|c|c|c|c|}
		\hline
		\textbf{Method} 
		& \textbf{\begin{tabular}[c]{@{}c@{}}Incorporates\\ temporal dimension\end{tabular}} 
		& \textbf{\begin{tabular}[c]{@{}c@{}}Smooth evolution\\ of a single vector\end{tabular}}
		& \textbf{\begin{tabular}[c]{@{}c@{}}Captures drifts in\\ long periods\end{tabular}} 		 
		& \textbf{\begin{tabular}[c]{@{}c@{}}Captures drifts in\\ short periods\end{tabular}} 
		& \textbf{\begin{tabular}[c]{@{}c@{}}Results are\\ low-dimensional\end{tabular}} \\ \hline
		tf-idf~\cite{sparck_jones_statistical_1972} & ~ & ~ & ~ & \checkmark & ~ \\ \hline
		word2vec~\cite{mikolov_distributed_2013} & ~ & ~ & ~ & ~ & \checkmark \\ \hline
		Dynamic Bernoulli embeddings~\cite{rudolph_dynamic_2018} & \checkmark & ~ & \checkmark & ~ & \checkmark \\ \hline
		Our proposed approach & \checkmark & \checkmark & \checkmark & \checkmark & ~ \\ \hline
		An aspiring approach & \checkmark & \checkmark & \checkmark & \checkmark & \checkmark \\ \hline				
	\end{tabular}
\end{table*}

In this work, we present a smooth and continuous time-reflective representation of temporal text data. To highlight the potential of the proposed time-reflective representation, we use the task of tracking the semantic evolution of words, which refers to identifying changes in the \textit{meaning} of words over time. The term \textit{meaning} in our work does not refer to the definition of a word as seen in the dictionary, but to the connotation inferred by studying the context of a word at different time stamps. In this paper, we use the phrase \textit{semantic evolution} and \textit{contextual evolution} interchangeably.

Figure~\ref{fig:ipad} (a) provides an example of the neighborhood of the word \textit{cloud} using a static representation. Figure~\ref{fig:ipad} (b) shows how the context words of the word \textit{cloud} evolved over time. Prominent context words of \textit{cloud} using a static representation are \textit{vapor}, \textit{rain}, \textit{Google Drive}, and \textit{Dropbox}. The history of how the neighboring words came into the context of \textit{cloud} is not apparent when using static embeddings. In the time-reflective representation (Figure~\ref{fig:ipad} (b)), it is evident that the terms \textit{Google Drive} and \textit{Dropbox} became more prominent in recent years. \textit{Context} is more time-dependent, as illustrated in Figure \ref{fig:ipad}. Our representation allows us to smoothly track the semantic evolution. While a static representation may provide the context of each word of a corpus as a set of nearest neighbors, the time-reflective representation has the ability to capture how the context of a word changes over time.

To study semantic evolution, the corresponding text corpus must be large and timestamped so that enough evidence can be gathered by a model to form a chain of concepts over time. News articles and scientific papers are good examples of timestamped text corpora. In news articles, a time-reflective representation can capture the evolution of a particular event over time. In scientific articles from a biomedical corpus, scientists can discover how a particular medical concept evolved in the past.

Semantic evolution is not yet quantifiable in the literature and cannot be easily modeled using data mining and machine learning algorithms. Moreover, the evolution of some words can be slow while for other words it can be fast. For example, the word \textit{husband} in early 14th century meant \textit{house-owner} and in the next several hundred years the meaning changed to \textit{a marital status}. In contrast, words like \textit{cloud}, \textit{apple}, and \textit{viral} have rapidly changed their meaning during the last two decades. We would like to have a temporal text representation that includes the following features:

\begin{enumerate}
	\item A temporal dimension that allows modeling evolution over time.
	\item Support for vector space modeling. That is, each word must have a vector in each timestamp so that the vectors can be tracked over time for each word.
	\item Continuous and smooth changes of a word vector over time, e.g., the vectors should not completely shift in space from one timestamp to the other. This property will allow usage of time series analysis algorithms.
	\item Ability to capture significant changes in the semantics of a word in a short period of time (e.g. between consecutive timestamps).
	\item Low dimensionality, i.e., the representation should be as compact as possible to make it suitable/feasible for use with big data analytics algorithms.
\end{enumerate}

Table~\ref{table:comparison} lists several vector space models in terms of their fulfillment of the five expected features of temporal text representations. There are two main types of vector space models for text data: co-occurrence- and embedding-based models. Word2vec~\cite{mikolov_distributed_2013} and dynamic Bernoulli embeddings~\cite{rudolph_dynamic_2018} are embedding-based methods, while tf-idf~\cite{sparck_jones_statistical_1972} and the proposed model are co-occurrence based models. As noted in Table~\ref{table:comparison}, this paper targets most of the expected properties of a time-reflective representation except for the low-dimensionality of the vectors. The aspiring approach that includes all the properties remains as a future work.  

\textbf{Pros and cons of co-occurrence-based models:}
These approaches are known to be accurate in terms of inferring the meaning of a given word based on the words that co-occur with the given word. Their disadvantage --- when used in tracking temporal semantic evolution --- is that the distribution of words that appear at every timestamp is usually sparse. As a result, it is possible that some words do not appear at all for a particular timestamp, making the representation discrete rather than continuous. A word might have a neighborhood that does not describe an accurate context of the word when the appearance of a word is infrequent in a particular timestamp. Another disadvantage of this approach is that it requires $O(mnt)$ storage, where $m$ is the number of documents, $n$ is the number of words, and $t$ is the number of timestamps in the corpus. 

\textbf{Pros and cons of embedding-based models:}
Word embeddings are a low-dimensional representation of a text corpus obtained using the co-occurrence information. This model is suitable for analyzing large text datasets due to its small footprint. Static embeddings require $d\times n$ storage where $d$ is the length of the vector of each word and $n$ is the number of unique words in the corpus, resulting in a $O(n)$ space requirement. If a time-reflective embeddings is created, the storage requirement will be $O(nt)$, given $t$ is the number of timestamps. One drawback of using word embeddings is that the vectors do not have interpretable values, unlike co-occurrence based approaches. The vectors provide a holistic information space in which nearest neighbors of each word represent the context of the word.
%However, in this work we show that for the task of temporal semantic similarity tracking, this method results in a ``too smooth'' representation that filters out sudden changes from one timestamp to the next. 

%\begin{itemize}
%	\item co-occurrence-based model: This representation is very accurate in terms of inferring the meaning of a word based on the word's context. Words that appear very frequently in the context of the word of interest provide meaning. The main disadvantage of this approach when using it in temporal semantic similarity tracking is that the distribution of words that appear at every timestamp is usually sparse, and thus, it is possible that some words do not appear at all for a particular timestamp, making this representation discontinuous. A word might have an inaccurate neighborhood, or even a random set of nearest neighbors when the appearance of a word is too sparse for a particular timestamp. Another disadvantage of this approach is that it requires $O(n^2)$ storage. 
%	\item embedding-based model: Word embeddings are low-dimensional representations of a corpus. The main advantage of this method is that it reduces the required storage to $O(n)$, making these vectors more suitable as inputs for highly-complex algorithms. However, in this work we show that for the task of temporal semantic similarity tracking, this method results in a ``too smooth'' representation that filters out sudden changes from one timestamp to the next.
%\end{itemize}

In this paper, we introduce a new co-occurrence-based time-reflective vector space model, which is diffusion-centric in the time dimension, referring to the change of the distributional patterns of a phenomenon over time. The model takes into account the co-occurrence of words within the same context for a particular timestamp, thus enabling the model to capture drifts in short time spans. We incorporate \textit{diffusion} of the word counts across timestamps to smooth the word vectors over time. The contributions of this paper are summarized as follows:
\begin{enumerate}
	\item We present a new diffusion-based method of generating co-occurrence-based time-reflective vector space models for temporal text corpora.
	%\item We introduce the neighborhood flatness metric which allows us to quantify the semantical change of a word over time.
	\item We introduce the \textit{neighborhood monotony metric}, which allows us to quantify the semantical or contextual change of a word over time.	
	\item We compare the new model with regular co-occurrence-based and dynamic-embedding-based vector space models.
\end{enumerate}

\section{Related Work}\label{sec:related}
Language has evolved over time, going through several syntactic and semantic diachronic (temporal) changes. This can be observed in words that disappear or are substituted by new words that better exemplify the contemporary meaning and intent of the expression~\cite{aitchison_language_2013, yule_study_2017}. There are many probabilistic approaches that aim to track temporal changes in word meaning and semantic evolution by transforming a text corpus into a latent time series model \cite{sagi_tracing_2011, radinsky_learning_2012, basile_analysing_2014, yogatama_dynamic_2014, frermann_bayesian_2016, tang_semantic_2016}. Mihalcea, et al.~\cite{mihalcea_word_2012} leverage Part of Speech (PoS) features from a word and its context to create a supervised model that predicts when that word was published. Other approaches represent a corpus as a graph, where each word is represented as a node and the edges between them are weighed based on contextual information~\cite{mitra_automatic_2015, kenter_ad_2015}. None of these approaches focus on finding a vector space model that accurately models the corpus for tracking semantic evolution.

\subsection{Dynamic topic models}
Several different versions of dynamic topic models have been developed. 
These models obtain the distribution of a word over latent topics, effectively trying to identify the changes in the patterns of word usage~\cite{blei_dynamic_2006, wang_topics_2006, wang_continuous_2008, hall_studying_2008, heyer_change_2009, wijaya_understanding_2011, naim_scalable_2017}. Dynamic topic models combined with clustering techniques can identify significant changes in the clusters over time~\cite{wijaya_understanding_2011}. However, these methods obtain the distribution of a word over topics, thus are limited in their application into finding the semantic changes of a word.

\subsection{Vector space models}
A vector space model is a tool that represents a text corpus in a continuous space. Vector space models can be classified into three classes: term-document, word-context, and pair-pattern~\cite{turney_frequency_2010}. The term-document matrix representation corresponds to the traditional tf-idf model~\cite{sparck_jones_statistical_1972} and its many different variations in terms of weighting and normalization~\cite{salton_term-weighting_1988, singhal_document_1996} at a document level. The word-context model is usually obtained at a window or sentence level. 
The pair-pattern matrices are obtained by identifying patterns (columns) for a pair of words (rows)~\cite{turney_frequency_2010}. These vector models can be leveraged to compute semantic neighborhoods by using cosine similarity. Hamilton, et al.~\cite{hamilton_cultural_2016} use a second-order vector which is obtained from the original vector plus the vectors of the neighborhood to compute the semantic similarity at each timestamp.

\subsection{Word embeddings}
Word embeddings are, in general, vector space models obtained by leveraging the distributional statistics of the words in a 
corpus. Thus, word embeddings over time can flexibly be used to compute the semantic similarity between words at different timestamps.  
Mikolov, et al.~\cite{mikolov_distributed_2013, mikolov_efficient_2013} introduced the word2vec model which obtains static embeddings, or low-dimensional representations of a corpus at a single point in time. Word2vec represents the foundation of most of the current state-of-the-art dynamic embeddings. Barkan~\cite{barkan_bayesian_2017} proposes a probabilistic version of the same algorithm using Bayesian inference. A different unsupervised learning approach is GloVe by Pennington, et al.~\cite{pennington_glove:_2014} in which word vectors are created using matrix factorization of a word-word co-ocurrence matrix.

The introduction of the temporal dimension in word embeddings has also been explored. Several authors have trained word embeddings at every timestamp in their corpus, and then used a method such as regression to artificially connect the embeddings over time~\cite{kim_temporal_2014, kulkarni_statistically_2015, hamilton_diachronic_2016, zhang_past_2016}. The main drawback of this approach in terms of the semantic evolution tracking task is that it requires having a very large number of documents at each timestamp so that word2vec (or a different static method) can obtain a high quality model, which is not always the case, specially with the introduction of new words at different points in time. 
Rosin, et al.~\cite{rosin_learning_2017} use a similar technique but introduce the supervised task of semantic relatedness,  which attempts to predict if two words are related at a certain point in time.

Dynamic embedding models have been introduced to overcome the problem of data sparsity and lack of continuity of the previous approach, by using joint embedding generation. Bamler and Mandt~\cite{bamler_dynamic_2017} leverage a probabilistic Bayesian version of word2vec to infer the embedding vectors at each timestamp, but use Kalman filtering to connect the embeddings over time. Yao, et al.~\cite{yao_dynamic_2018} propose an approach that uses the pointwise mutual information (PMI) matrix instead of word2vec. To obtain the dynamic embeddings, the PMI matrix is factorized iteratively at each timestamp, and alignment is enforced through regularization. Finally, Rudolph and Blei~\cite{rudolph_dynamic_2018} use Kalman filtering like Bamler and Mandt, but Rudolph and Blei use a non-Bayesian approach based on \textit{exponential family embeddings}. According to Bamler and Mandt, using such a non-Bayesian approach makes the model more sensitive to noise for sparse data. 

% \subsection{Embedding/semantic change evaluation}
% The vast amount of work on word embeddings and semantic change analysis has resulted in a significant interest on evaluating the quality of the results~\cite{schnabel_evaluation_2015, gladkova_intrinsic_2016, hellrich_bad_2016}, as well as visualizing the semantic evolution of words in a comprehensible, useful way~\cite{jatowt_framework_2014}. Schnabel, et al. introduce a \textit{coherence} metric that is evaluated by showing a group of three words semantically related to a word at certain point in time to a human evaluator, as well as an unrelated word. The expectation is that the evaluator will be able to easily identify the non-related word~\cite{schnabel_evaluation_2015}. Hellrich and Hahn~\cite{hellrich_bad_2016} show that ``reliable'' algorithms based on word2vec usually provide inconsistent word neighborhoods. Dubossarsky, et al.~\cite{dubossarsky_outta_2017} argue that the evidence of semantic laws given by recent literature is not enough and that some of those laws are simply an artifact of the characteristics of the data.

\section{Problem Description}\label{sec:problem}
This paper focuses exclusively in timestamped text corpora, such as collections of news articles or scientific publications. 
Let $\mathcal{D} = \{{d}_1, {d}_2, \ldots, {d}_{|\mathcal{D}|}\}$ be a corpus of $|\mathcal{D}|$ text documents and $\mathcal{W} = \{w_1, w_2, \ldots, w_{|\mathcal{W}|}\}$ be the set of $|\mathcal{W}|$ noun phrases extracted from the text corpus $\mathcal{D}$. 
Each document $d$ contains noun phrases from the vocabulary ($\mathcal{W}_d \subset \mathcal{W}$) in the same order as they appear in the original document of ${d}$.
Every document ${d} \in \mathcal{D}$ is labeled with a timestamp $t_{d} \in \mathcal{T}$, where $\mathcal{T}$ is the ordered set of timestamps.

\subsection{Expected outcome} \label{sec:problem:outcome}
The goal of this paper is to obtain a time-reflective vector space model from corpus $\mathcal{D}$. Thus, for every timestamp $t \in \mathcal{T}$, we seek to obtain a vector representation of every word $w \in \mathcal{W}$. Ideally, the vectors of words that appear frequently in the same \textit{context} of word \textit{A} should be \textit{spatially closer} to the vector of word \textit{A}, than those that appear less often. 

\section{Methodology for Diffusion-based Semantic Tracking}\label{sec:methods}
In this work we focus on a specific application of vector space models --- 
tracking semantic or contextual evolution. We focus in such a task because it allows the user to determine the meaning of a particular word at different times. The context of a word can be retrieved from the nearest neighbors of a word vector. 
Our approach consists of two main components: 
(1) temporal diffusion of words and 
(2) temporal neighborhood retrieval.

\subsection{Temporal diffusion}
The distribution of the frequency of a word over time can be severely irregular, in particular for words or noun phrases that suddenly appear at a particular timestamp, or that are used sporadically. Furthermore, based on \textit{diffusion theory}~\cite{angulo_concepts_1980}, which refers to the change of the distributional patterns of a phenomenon over time, we assume that the meaning of a word, and consequently its vector representation, will diffuse over time.

To smooth the word vectors over time, we assume that every document is present 
in every timestamp. However, those timestamps closer to when the document was originally published have a higher probability. We use a Gaussian filter to \textit{diffuse} the contribution of the document smoothly before and after the publication date of the document. 
The filter uses a sliding window, going from the first to the last timestamp. 
%Figure~\ref{fig:gauss} illustrates the filter being used. 
The contribution of a document $d$ increases the closer its timestamp $t_d$ is to the current timestamp $t$. The tf-idf weight of a word can be modified at each timestamp with Equation~\ref{eq:filter}.
\begin{multline}\label{eq:filter}
	\hat{w}(w, d, t_d, t, \sigma)=\left(\frac{1}{\sqrt{2\pi\sigma^2}}e^{-\frac{(t_d-t)^2}{2\sigma^2}}\right) \cdot\\
	\left(\frac{(1 + \log(f_{w,d})(\log{\frac{|\mathcal{D}|}{\delta_w}})}{\sum_{w' \in \mathcal{W}_d }{\left((1 + \log(f_{w',d})(\log{\frac{|\mathcal{D}|}{\delta_{w'}}})\right)^2}}\right),
\end{multline}
where $\hat{w}$ is the weighted value at timestamp $t$ for the noun phrase $w \in \mathcal{W}$ in document $d \in \mathcal{D}$, which was published at timestamp $t_d$. The term $f_{w,d}$ represents the term frequency of noun phrase $w$ in document $d$, $\delta_w$ is the number of documents that contain noun phrase $w$, and $\mathcal{W}_d$ is the set of noun phrases that appear in document $d$.
$\sigma$ represents the standard deviation of the Gaussian distribution, and is set by the user. A large value of $\sigma$ means that the diffusion of concepts will be slow over time. A small standard deviation will allow capturing short-term changes in meaning, but makes the model more susceptible to noise.

%\begin{figure}
%	\includegraphics[width=\linewidth]{gaussian}
%	\caption{Gaussian distribution. \label{fig:gauss}}
%	\centering
%\end{figure}

\subsection{Computing the nearest neighbors of a word in a particular timestamp}
To track the semantic evolution of a given word, we perform an analysis from one year to another and construct a set of $k$-nearest neighbors for each year. Analyzing the changes in a word's neighborhood in consecutive years helps us understand the semantic change of a word over time. 

The \textit{context} of a word conveys the meaning and intent of the word. Selecting a context that filters out unimportant or irrelevant words can improve significantly the quality of the retrieved neighborhoods. 
In this work, we filter out every word occurrence that is not within the context of the word under analysis, and we evaluate the following contexts:
\begin{itemize}
	\item \textbf{document-level context.} Every word in document $d$ is part of the context of the other words in the same document.
	\item \textbf{window-level context.} Only the words that are within a window of a particular size from the word of interest are part of its context. For example, if the window size is 2, then the two words before and the two words after the word of interest will belong to its context.
	% \item \textbf{relation-level context.} Using a relation-extraction algorithm, we can choose a context that includes only the words that have some kind of relation with the word of interest.
\end{itemize}

The neighborhood retrieval task is divided in three main subtasks. First, we find all the occurrences of words that belong to the context of the word of interest. Next, we set all of the tf-idf vector entries that do not belong to the context to 0. Finally, we compute the cosine similarity between the resulting vectors for every timestamp. The neighborhood is formed by the terms that have the highest values of cosine similarity at each timestamp.

%The cosine distance is defined by Equation~\ref{eq:cosine}, where $\mathbf{W}$ represent word vectors, $\cdot$ represents the dot-product and $||\cdot||$ represents the $l_2$-norm. The vectors that have the smallest distance between each other are considered to be the nearest neighbors.
% \subsubsection{Cosine distance}
%\begin{equation}\label{eq:cosine}
%	\text{cosine}(\mathbf{W}_i, \mathbf{W}_j)=1-\frac{\mathbf{W}_i \cdot \mathbf{W}_j}{||\mathbf{W}_i||\, ||\mathbf{W}_j||}
%\end{equation}

%\subsection{Evaluation of semantic similarity tracking}
\subsection{Evaluation of semantic evolution}
The meaning of a word tends to evolve and change over time. Usually, we can infer the meaning of a word by looking at the context in which it appears most frequently. Thus, we assume that identifying significant changes in the context of a word will help us determine if its meaning is also changing. In this paper, we introduce the concept of \textit{neighborhood monotony} to evaluate, quantitatively, how much the context of a word changes over time, based on the word's neighborhood over consecutive timestamps.
This metric helps in evaluating semantic evolution by allowing to estimate the degree to which the semantical or contextual meaning remains steady.
We name such steady-ness of the meaning of a word throughout a timeline \textit{neighborhood monotony}. The neighborhood monotony, $\bar{F}$, is computed by taking the average of Jaccard similarities~\cite{rogers_computer_1960} between the neighborhoods of a word in every consecutive pair of timestamps.
%We measure steadyness computing the Jaccard similarity between the neighborhoods of consecutive timestamps for a single word. 
Equation~\ref{eq:monotony} formulates the average neighborhood monotony of word $w$.

\begin{equation}\label{eq:monotony}
	\bar{F}(M, w, k) =\frac{1}{|T|-1}\sum_{i=0}^{|T|-1}{\frac{M(t_i, w, k)\cap M(t_{i+1}, w, k)}{M(t_i, w, k)\cup M(t_{i+1}, w, k)}},
\end{equation}
where $M(t_i, w, k)$ is the neighborhood of size $k$ for word $w$ at timestamp $t_i$.
%$w$ is the word of interest, 
%$k$ is the size of the neighborhood, and $t_i$ is the $i$\textsuperscript{th} timestamp.

If the $k$-neighborhood of a word remains unchanged across all timestamps, the word is considered to exhibit a completely \textit{monotonous} evolution, and the average neighborhood monotony will be 1.0. If the neighborhood changes completely for every pair of consecutive timestamps, then the average neighborhood monotony will become 0.0. 

Additionally, we introduce the concept of \textit{minimum} and \textit{absolute} neighborhood monotony. \textit{Minimum neighborhood monotony} refers to the pair of consecutive timestamps where the Jaccard similarity between neighborhoods is lowest. It identifies significant changes in the meaning of a word in a single point in time. \textit{Absolute neighborhood monotony} refers to the Jaccard similarity between the neighborhoods of the first and last timestamps. It helps identify very monotonous concepts that did not evolve over time.

%Additionally, we introduce the concept of \textit{minimum} and \textit{absolute} neighborhood flatness. Minimum neighborhood flatness refers to the pair of consecutive timestamps where the Jaccard similarity between neighborhoods is lowest, and it identifies significant changes in the meaning of a word in a single point in time, in particular for words that have a high average neighborhood flatness. Absolute neighborhood flatness refers to the Jaccard similarity between the neighborhoods of the first and last timestamps, and helps identify very flat concepts that did not evolve over time.
% \subsubsection{Soergel distance}
% The Soergel distance between a pair of documents $d_i$ and $d_j$, defined in Eq.~\ref{eq:soergel}.

% % Soergel distance (3)
% \begin{equation} \label{eq:soergel}
% \text{soergel}(\mathbf{W}_i, \mathbf{W}_j) = \frac{\sum_{k=0}^{|\mathbf{W}|}\left|\mathbf{W}_{ik}-\mathbf{W}_{jk}\right|}
% {\sum_{k=0}^{|\mathbf{W}|}\max\left(\mathbf{W}_{ik},\mathbf{W}_{jk}\right)}
% \end{equation}
% \noindent where $|\mathbf{W}|$ is the length of a single word vector. 

\section{Experimental Results}\label{sec:experiments}
For our experiments, we use a corpus of 613,545 PubMed abstracts~\cite{bethesda_pubmed_1946} related to the query ``anticancer agents'', published between January 1998 and April 2018. We preprocessed each abstract by extracting its noun-phrases using the SpaCy Python library~\cite{honnibal_spacy_2017}. The corpus contains 15,988,890 unique noun-phrases. We selected the top-10,000 noun-phrases based on frequency. We evaluate our method by comparing its performance with that of a regular tf-idf model and the state-of-the-art dynamic Bernoulli embedding model~\cite{rudolph_dynamic_2018}.
%We briefly describe the particular flavors and parameters used for these methods next.
%For our experiments, we use a corpus of 400,842 New York Times articles published between January 2000 and June 2016 on the U.S. and World news sections. The corpus contains 3,320,886 unique entities, but we select only the top-25,000 entities based on frequency. We seek to answer the following questions.

%\subsubsection{Tf-idf model}
In the tf-idf model, each document $d \in \mathcal{D}$ is represented as an $|\mathcal{W}|$-length sparse vector, which contains the normalized tf-idf weights of each noun phrase $w\in \mathcal{W}$. To compute these weights we use tf-idf with cosine normalization~\cite{hossain_storytelling_2012}.
% :
% \begin{equation} \label{eq:tfidfnormalized}
% 	\text{tf-idf}(w, d) = \frac{(1 + \log(f_{w,d})(\log{\frac{|\mathcal{D}|}{\delta_w}})}{\sum_{w' \in \mathcal{W}_d }{\left((1 + \log(f_{w',d})(\log{\frac{|\mathcal{D}|}{\delta_{w'}}})\right)^2}},
% \end{equation}
% where $f_{w,d}$ represents the term frequency of noun phrase $w \in \mathcal{W}$ in document $d \in \mathcal{D}$, $\delta_w$ is the number of documents that contain noun phrase $w$, and $\mathcal{W}_d$ is the set of noun phrases that appear in document $d$.

For dynamic Bernoulli embeddings, we use the code provided by Rudolph and Blei~\cite{rudolph_dynamic_2018}, and apply the same parameters that they described. However, the optimal configuration for the results presented in \cite{rudolph_dynamic_2018} is not provided in the paper. Therefore, we performed a thorough exploration of different configurations and selected the model that provides the best test log likelihood, which was $-2.71$. The parameters used in our work are:
\begin{itemize}
	\item embedding size: 100
	\item window size: 2 (i.e. 2 words on each side)
	\item negative samples: 20
	\item number of batches: $100000 \in [10000, 100000]$
\end{itemize}
% For NVD: d_100_4_20_10000_10_0.01/variational0.dat

%\subsubsection{Temporal tf-idf}
For our own model (Equation \ref{eq:filter}), we set the temporal diffusion parameters to the following values:
\begin{itemize}
	\item standard deviation of the diffusion over time: 1.0
	\item context type: window
	\item window size: 2 (i.e. 2 words on each side)
\end{itemize}

%\subsection*{Research questions}
In this section, we seek to answer the following questions.
\begin{enumerate}
	\item How well does our algorithm track the evolution of specific medical terms? (Section~\ref{sec:exp:example})
	\item How does our method compare qualitatively to other methods in terms of tracking semantic evolution? (Section~\ref{sec:exp:qual})
	\item How sensitive is our vector space model to changes in the hyperparameters? (Section~\ref{sec:exp:sensitivity})
	\item How does our method compare quantitatively to other methods such as regular tf-idf and dynamic embeddings in terms of neighborhood monotony? (Section~\ref{sec:exp:quant})
\end{enumerate}

\subsection{Case study} \label{sec:exp:example}
In this section, we present a study in tracking semantic evolution using our algorithm for two medical terms: \textit{leukemia} and \textit{gastric cancer}. We limit our analysis to the top-16 nearest neighbors for each timestamp, from which we present the most interesting words that show significant changes over time. The results are promising because, as we describe next, the neighborhoods help explain the context or contemporary state-of-the-art related to a particular term.

\subsubsection{Evolution of ``leukemia'' neighborhood}

\begin{figure*}
	\centering
	\includegraphics[width=0.8\textwidth]{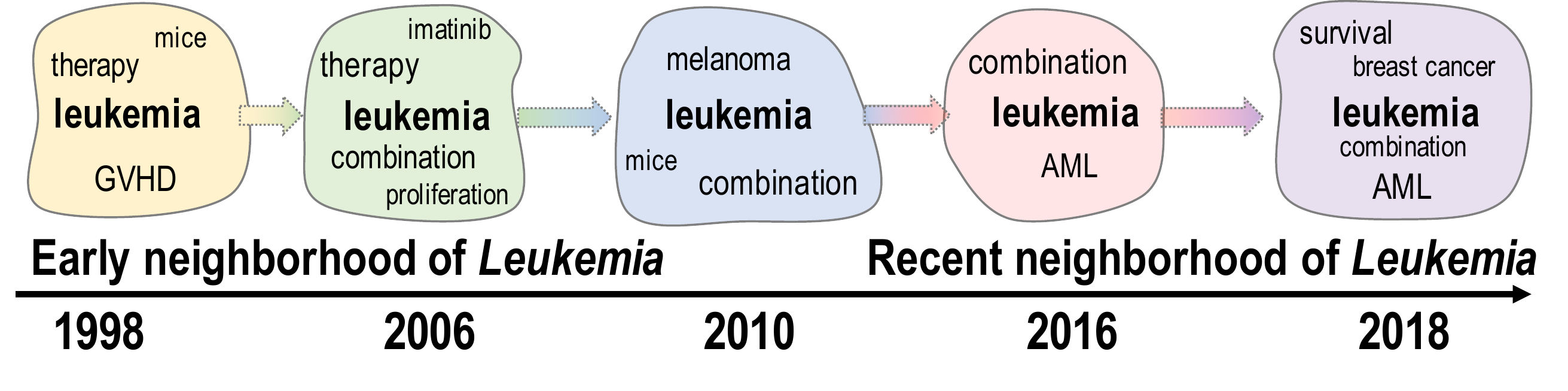}
	\caption{Neighborhood of the word \textit{leukemia} in biomedical abstracts over time, derived from our time-reflective model.}
	\label{fig:leukemia_evolution}
\end{figure*}

%Figure~\ref{fig:leukemia_evolution} illustrates the evolution of the term \textit{leukemia} over the years captured computing the semantic similarity between the vectors obtained using our representation from a medical abstract database. As we explain in Section~\ref{sec:exp:example}, our approach is able to capture significant changes such as the introduction of the term \textit{breast cancer} in the neighborhood, which indicates a change in meaning of the word \textit{leukemia}, since in this case \textit{leukemia} is referring to the \textit{Bovine Leukemia Virus} being correlated with breast cancer~\cite{baltzell_bovine_2018, cuesta_can_2018}, instead of the usual meaning of human blood cancer.

Figure~\ref{fig:leukemia_evolution} shows the evolution of the neighborhood of the term \textit{leukemia} in medical abstracts over the last twenty years. The early neighborhood includes the term \textit{GVHD}, which is a disease frequently related in the literature to bone marrow transplantation, and has been identified as the most important issue in the treatment of relapsing leukemia patients~\cite{min_effect_2001}.

\textit{Imatinib} is a drug approved by the FDA as a treatment for leukemia in 2001~\cite{dagher_approval_2002}. Around 2006, several articles were published that included long term studies of the effects and success rates of \textit{Imatinib}~\cite{druker_five-year_2006, obrien_imatinib_2003, kantarjian_hematologic_2002}, as well as several experiments that \textit{combine} this drug with some other treatments~\cite{yanada_high_2006}. This could also explain the appearance of the word \textit{combination} in the neighborhood starting in 2006. 

From the observed results, we can infer that \textit{combining} two or more treatments to try to cure leukemia has been the trend since 2006~\cite{fischer_bendamustine_2012, u.s._food_and_drug_administration_fda_2018}. There has also been an important recent effort on curing \textit{Acute Myeloid Leukemia} (\textit{AML}), which is a very common, fast-growing, and known to be deadly cancer~\cite{national_cancer_institute_adult_nodate}. In 2018, the FDA approved a combination treatment targeting \textit{AML}~\cite{u.s._food_and_drug_administration_fda_2018}. 

Finally, in 2018, several articles were published exploring the possibility that the presence of Bovine \textit{Leukemia} Virus (BLV) in humans can significantly increase the risk of \textit{breast cancer}~\cite{baltzell_bovine_2018, cuesta_can_2018}. In this case, we can infer that the word \textit{leukemia} is more closely related to the bovine virus associated \textit{breast cancer} and not to the commonly known human blood cancer. This example shows the potential of using our approach to identify multiple meanings for the same word. 

\subsubsection{Evolution of ``gastric cancer'' neighborhood}
\begin{figure*}
	\centering 		\includegraphics[width=0.8\textwidth]{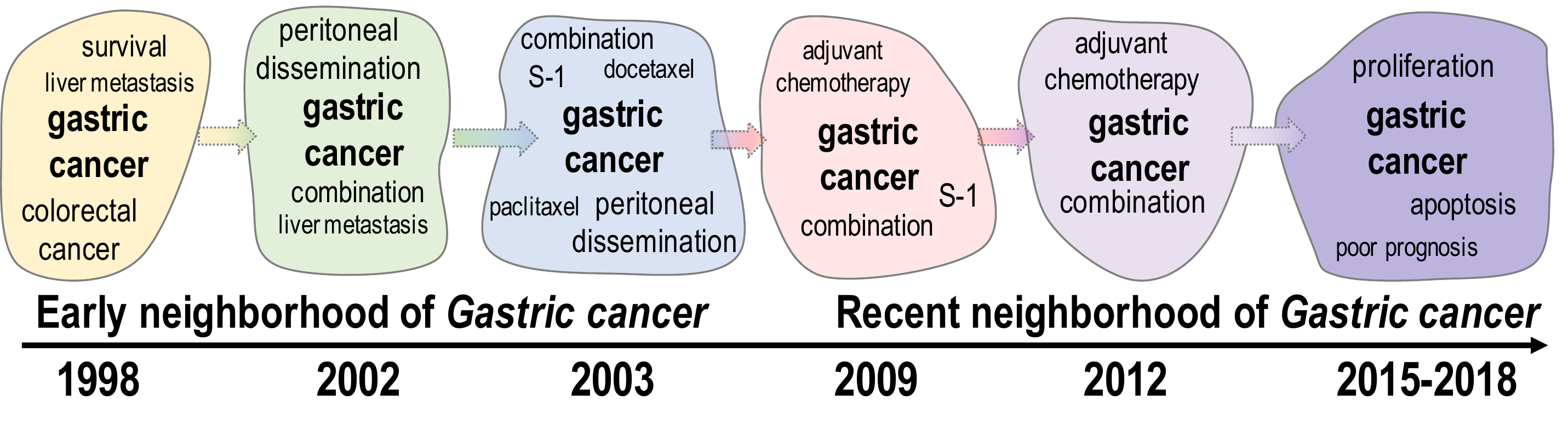}
	\caption{Neighborhood of the term \textit{gastric cancer} over time, computed using our time-reflective model.}
	\label{fig:gastric_cancer_evolution}
\end{figure*}
Figure~\ref{fig:gastric_cancer_evolution} presents the evolution of \textit{gastric cancer} for the last two decades. 
Between 1997 and 2003, many studies of \textit{liver metastases} caused by \textit{gastric cancer} were reported~\cite{liao_hepatectomy_2017}, which could explain its appearance in the early neighborhood of \textit{gastric cancer}. In the period of 2000 to 2004, there were several findings related to the role of integrins in the\textit{peritoneal dissemination} of gastric cancer cells, which is relevant because there is no effective treatment for \textit{gastric cancer} after this cancer stage is reached~\cite{kanda_molecular_2016}. 

To treat early stages of \textit{gastric cancer}, several different \textit{combinations} of treatments have been studied. Currently, however, there is no single treatment that is considered the most effective~\cite{american_cancer_society_treatment_2017, american_cancer_society_chemotherapy_2017}. \textit{Docetaxel}, \textit{placitaxel} and \textit{S-1} are drugs that have been used to treat gastric cancer, either separately or as a \textit{combination}~\cite{koizumi_phase_2003, mochiki_phase_2006,
yamaguchi_phase_2006}. Many of the publications between 2003 and 2005 about these drugs are phase 1 or 2 studies, which means that the drugs were in the early stages of clinical research at that time~\cite{u.s._food_and_drug_administration_drug_nodate}. Thus, we can infer that the novelty of these drugs caused a peak in the interest of the clinical research community during this period.

In 2007, \textit{S-1} was approved as an effective option for \textit{adjuvant chemotherapy} for patients with resected gastric cancer. Thus, it is possible that the term \textit{adjuvant chemotherapy} appeared within the neighborhood of \textit{gastric cancer} because many published studies evaluated its effectiveness~\cite{kilic_current_2016}.

Recently, there has been a significant interest on inducing \textit{apoptosis} in gastric cancer cells. \textit{Apoptosis} refers to ``auto-destruction''~\cite{xu_sp1-induced_2015} of the harmful cells. There has also been contemporary interest on identifying substances or conditions that suppress \textit{apoptosis}~\cite{jenkins_inflammasome_2017}. Finally, the terms \textit{proliferation} and \textit{poor prognosis} seem to be related to \textit{gastric cancer} because of many recent publications that correlate substances with the proliferation or suppression of gastric cancer cells~\cite{zhang_fam196b_2017, wang_downregulation_2018, zhu_dandelion_2017
}, as well as with indicators of a poor prognosis or outcome~\cite{zhang_high_2018, jiang_overexpression_2018}.

\subsection{Qualitative evaluation} \label{sec:exp:qual}
In this experiment, we use the task of tracking semantic evolution to illustrate the advantages of using our model. We compare the neighborhoods over time obtained using our method with those obtained using the tf-idf model and the dynamic-embedding-based model. For this experiment, we selected the top-16 words such that their tf-idf-based vectors have the highest cosine similarity at any point in time with the tf-idf vector for the word of interest. Next, we obtained the evolving trends for these 16 words using the dynamic Bernoulli embeddings and the vectors generated by our time-reflective model (Eq. \ref{eq:filter}). 

\begin{figure}
	\centering
	\subfloat[Tf-idf with cosine normalization
	%Tf-idf (Equation \ref{eq:tfidfnormalized})
	\label{subfig:leukemia_tfidf}]{
		\centering
		\includegraphics[width=0.94\columnwidth]{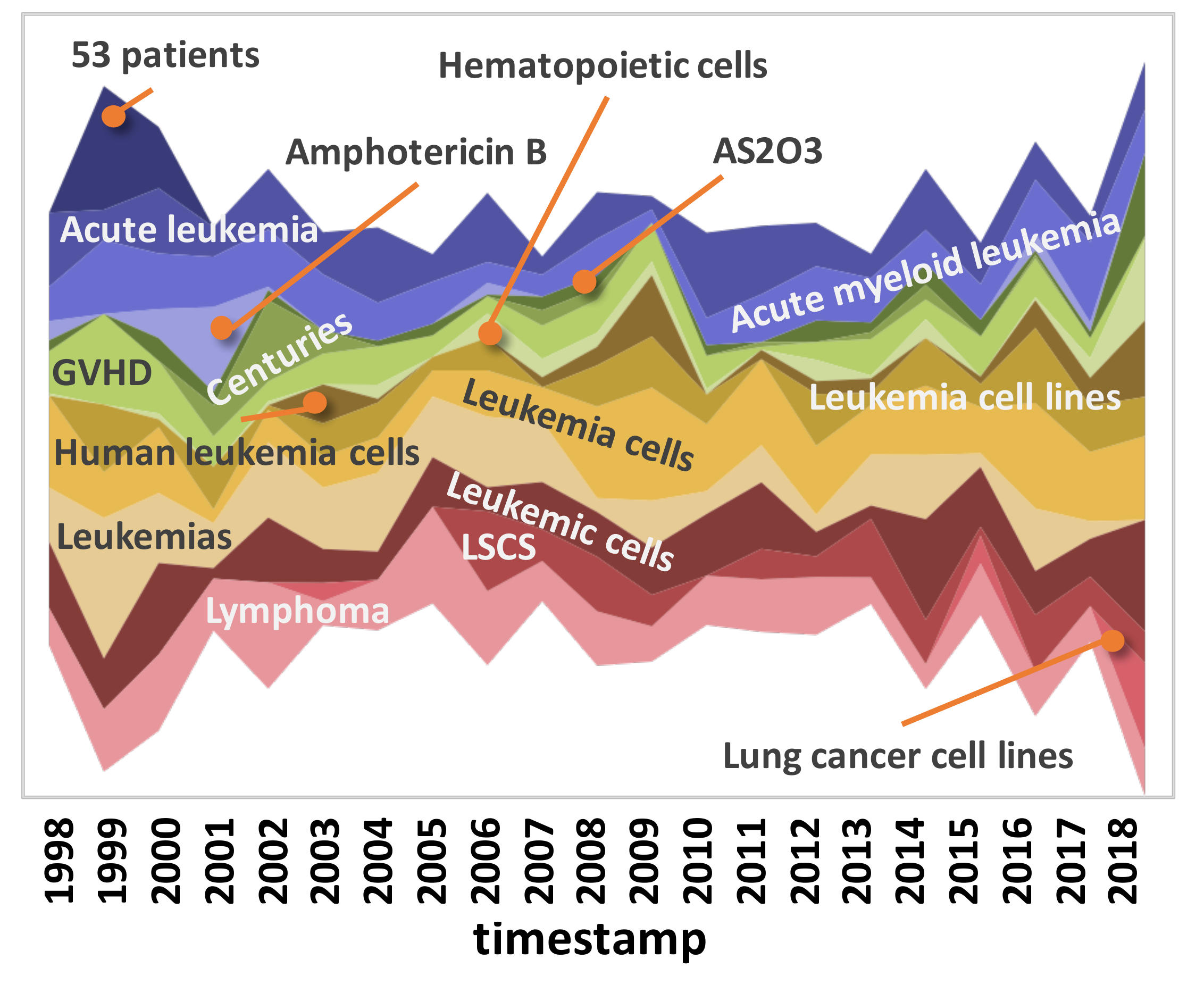}
	}
	\\
	\subfloat[Our time-reflective model (Equation \ref{eq:filter})\label{subfig:leukemia_temporal_tfidf}]{
		\centering
		\includegraphics[width=0.94\columnwidth]{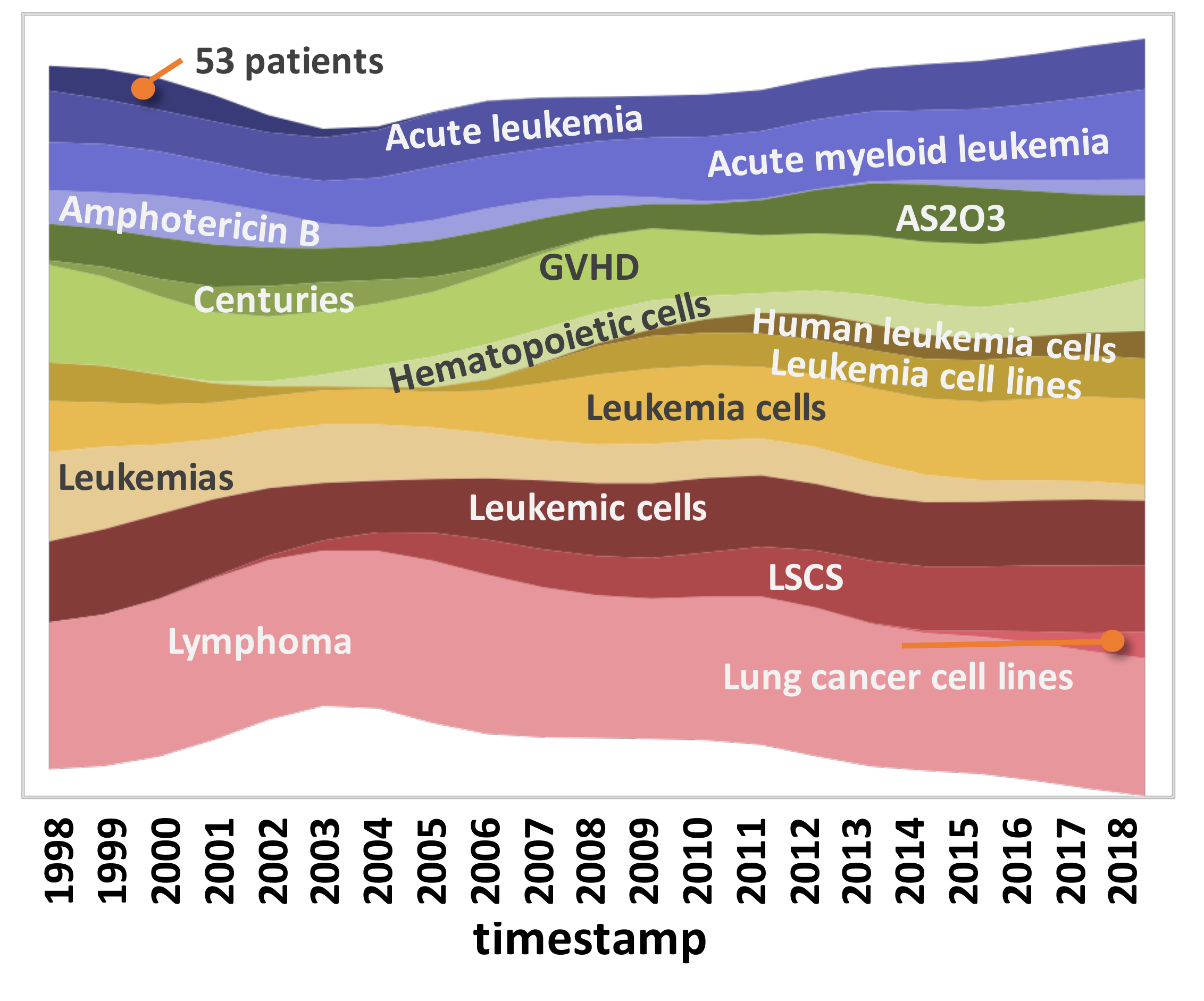}
	}
	\\
	\subfloat[Dynamic Bernoulli embeddings\label{subfig:leukemia_bernoulli}]{
		\centering
		\includegraphics[width=0.94\columnwidth]{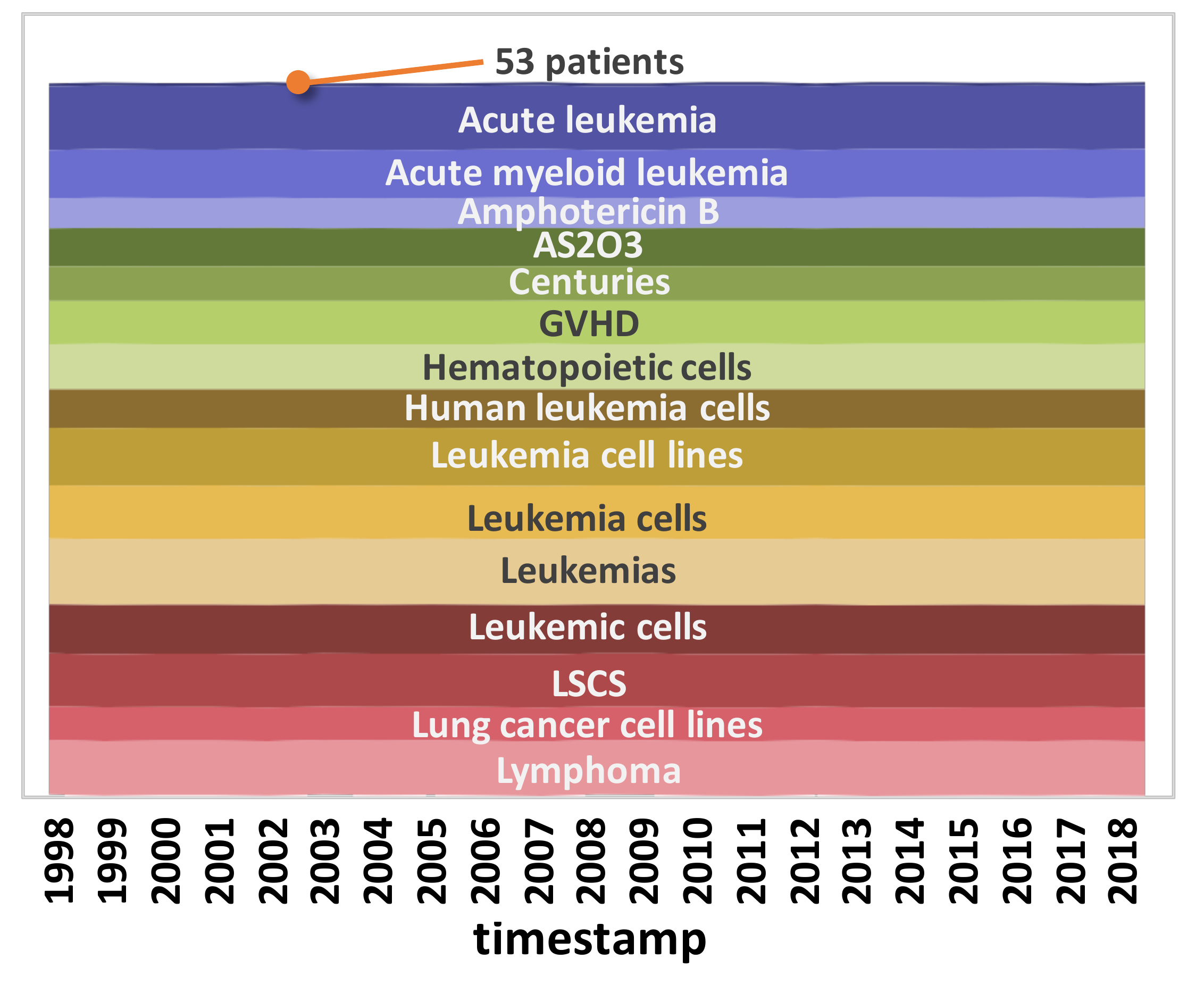}
	}
	\caption{Evolution of \textit{leukemia} using different vector representations.
		}
	\label{fig:leukemia}
\end{figure}

Fig.~\ref{fig:leukemia} shows the evolution of the neighborhood of the word \textit{leukemia} at different points in time: for (a) the tf-idf method, (b) our time-reflective method (Equation \ref{eq:filter}), and (c) the dynamic Bernoulli embeddings method. The height of a stream at a particular point represents the cosine similarity of that word vector with the vector for \textit{leukemia}. It is quite noticeable that the tf-idf method provides a noisy signal, while the dynamic Bernoulli method produces almost ``stale'' results. Clearly, none of the noticeable variations from the tf-idf model are captured. In contrast, our method captures the most noticeable variations from the tf-idf methods, e.g. for the noun phrases \textit{1000 mg}, \textit{lscs}, \textit{lung cancer cell lines}, and \textit{centuries}. Furthermore, we can observe that the trends are clearer in our method because the noise of the original data is filtered out.

\begin{figure}
	\centering
	\includegraphics[width=0.8\columnwidth]{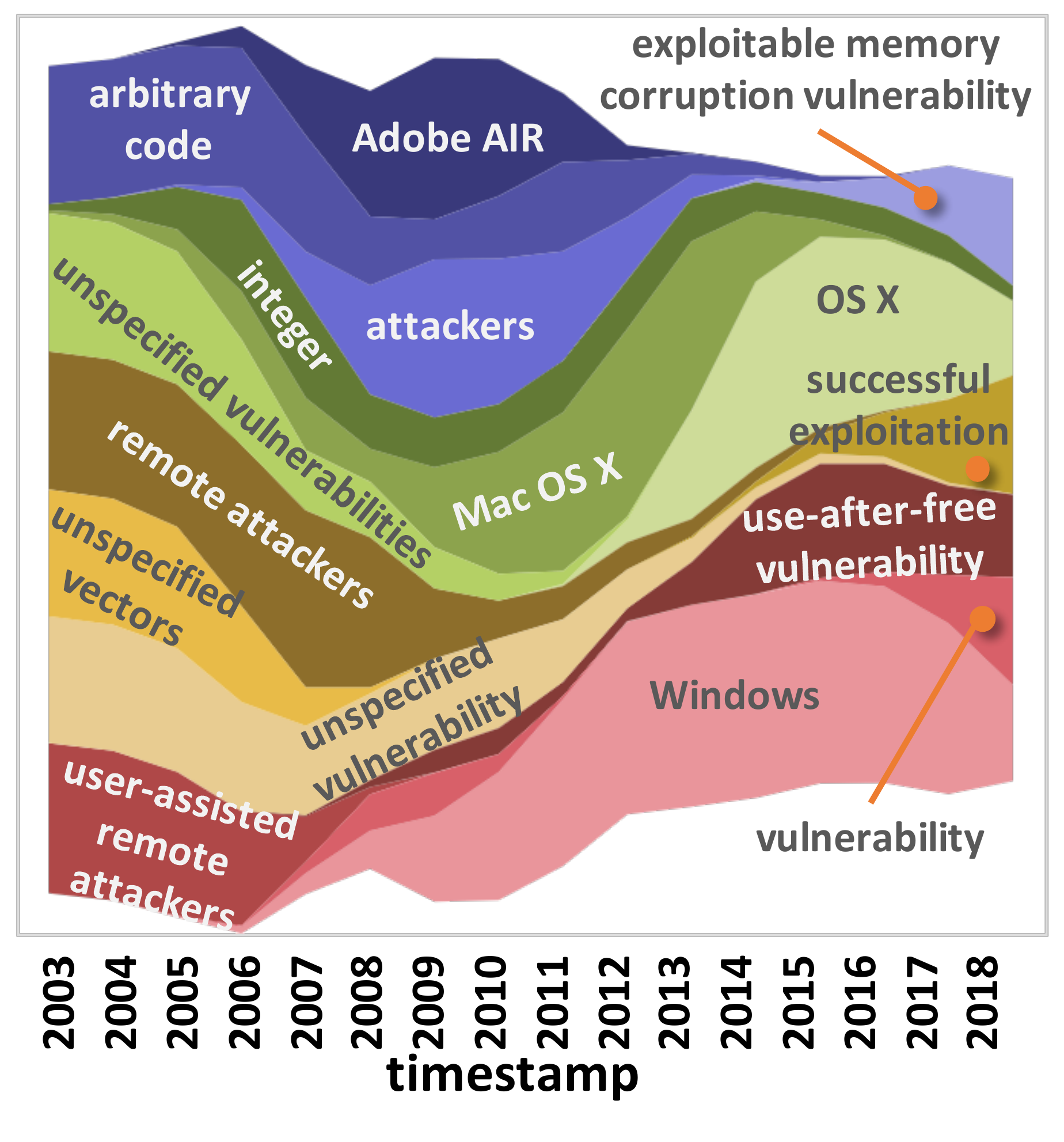}
	\caption{Evolution of \textit{Adobe flash player} using our time-reflective model (Equation~\ref{eq:filter}).
		}
	\label{fig:flash}
\end{figure}

We performed another experiment using data from the National Vulnerability Database~\cite{standards_national_nodate} to verify that our approach could be extended to other datasets. Figure~\ref{fig:flash} shows the evolution of the term \textit{Adobe flash player} from 2003 to 2018. In the figure we can identify two interesting trends: in the past, \textit{user-assisted remote attackers} and \textit{arbitrary code} where prominent, while recently terms such as \textit{exploitable memory corruption vulnerability} (with \textit{successful exploitation}) and \textit{user-after-free vulnerability} are more related to \textit{Adobe Flash}. The results for the dynamic Bernoulli embeddings and tf-idf are similar to those shown in Fig.~\ref{subfig:leukemia_tfidf} and Fig.~\ref{subfig:leukemia_bernoulli}.

Our webpage\footnote{https://sites.google.com/view/tracking-evolution/home} presents additional plots for \textit{Adobe flash} as well as other case studies from the PubMed dataset. The first one is for the term \textit{colon cancer} and the second one focuses on the term \textit{diffuse large B-cell lymphoma (DLBCL)}. Both of these additional case studies demonstrate that our time-reflective tracking of semantic evolution smoothly captures changes in the meanings of words.

\subsection{Sensitivity analysis} \label{sec:exp:sensitivity}
In this experiment, we evaluate the effect of performing a sweep of different values for (a) the standard deviation and (b) the word context on our model using our neighborhood monotony metric (Equation \ref{eq:monotony}). Ideally, we would like to capture a pattern that is not too monotonous (i.e. not close to 1.0), and not too unstable (i.e. not close to 0.0). We experimented using both document-level context and window-level context. The window-level context was varied between window sizes of 1 to 4. We used multiple standard deviations in our model (Equation \ref{eq:filter}): 0.5, 1.0, 2.0, 3.0, 5.0.

%The parameters and values that were evaluated in this experiment are the followings:
%\begin{itemize}
%	\item Context 
%	\begin{itemize}
%		\item Document level
%		\item Window level (window size $\in [1, 2, 3, 4]$)
%	\end{itemize}
%	\item Standard deviation of temporal diffusion
%	\begin{itemize}
%		\item $\sigma \in [0.5, 1.0, 2.0, 3.0, 5.0]$
%	\end{itemize}
%\end{itemize}
\begin{figure*}
	\centering
	\includegraphics[width=\textwidth]{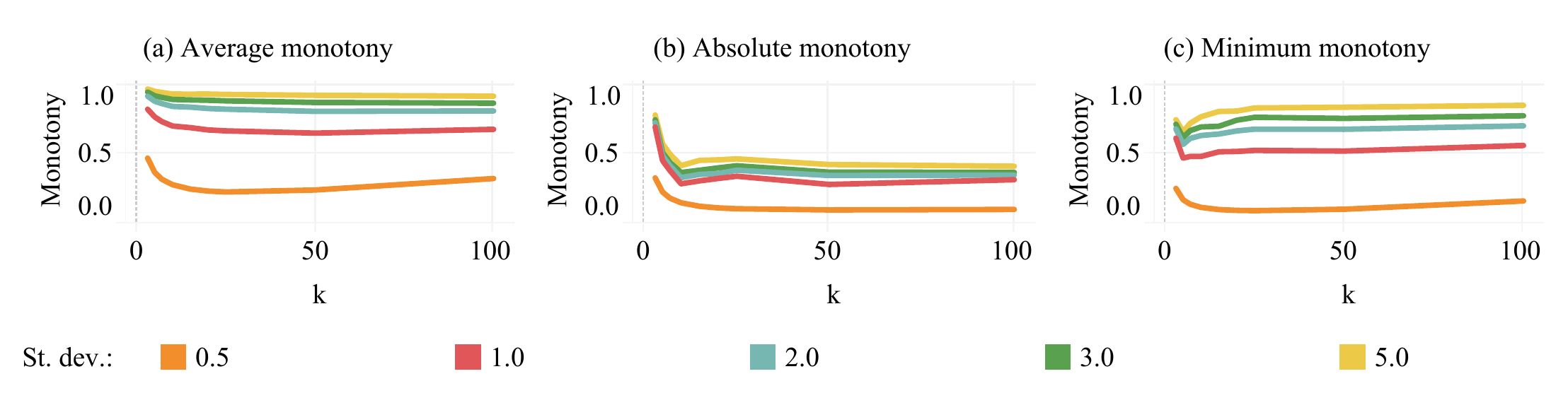}
	\caption{Average(a), absolute(b) and minimum(c) neighborhood monotony for different values of standard deviation.}
	\label{fig:stdev_flatness}
\end{figure*}

\begin{figure*}
	\centering	
	\includegraphics[width=\textwidth]{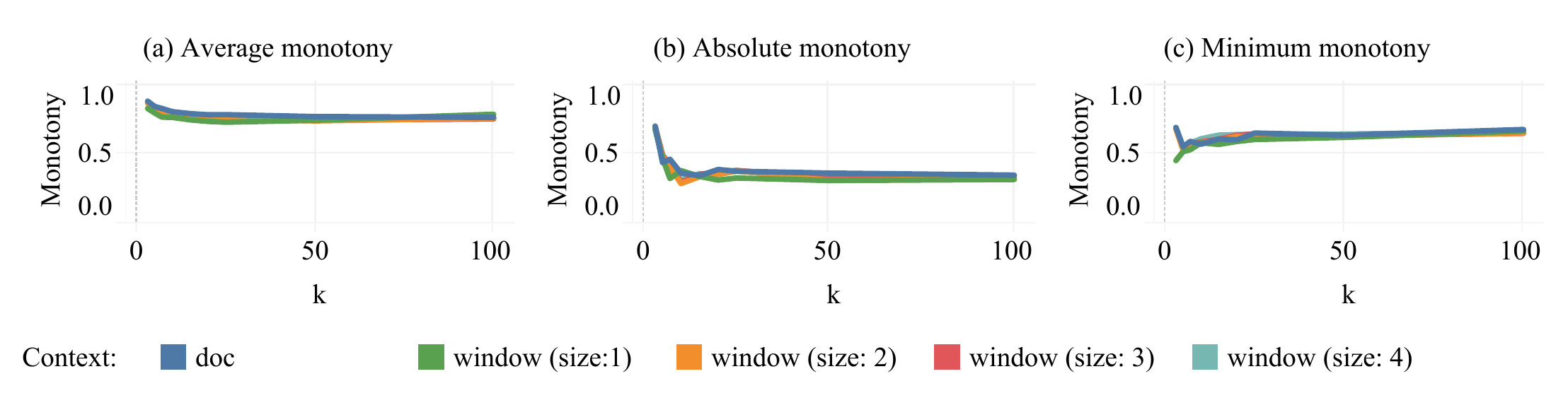}
	\caption{Average(a), absolute(b) and minimum(c) neighborhood monotony for different contexts.}
	\label{fig:context_flatness}
\end{figure*}

Figure~\ref{fig:stdev_flatness} shows how the neighborhood monotony metric reacts with varying neighborhood sizes ($k$) and varying standard deviation ($\sigma$). As we increase the value of the standard deviation, the neighborhood monotony increases significantly. This is to be expected because increasing the standard deviation means that we are increasing the contribution of documents from distant timestamps to the current timestamp. Based on the observed results, we select the optimal standard deviation value of 1.0, since this results in an average neighborhood monotony level that is not too monotonous, and not too unstable.

Figure~\ref{fig:context_flatness} shows how the different metrics of neighborhood monotony change for different neighborhood sizes ($k$) while varying the context of the selected representation. As we use different contexts, the neighborhood monotony values do not present a significant variation. We believe that this behavior results from the idea that words that are closer to the word of interest are more relevant and often appear together in many different documents. This means that the immediate context of a word will probably be part of its nearest-neighbor set. Thus, increasing the size of the window does not improve significantly the detection of changes in the neighborhood. Based on these results, we decided to use a window size of two words on each side of the center word, since this is also the value used by Rudolph and Blei~\cite{rudolph_dynamic_2018}.

\subsection{Quantitative comparison with other methods} \label{sec:exp:quant}
A quantitative evaluation is significantly difficult because there is no method that can be considered a ``ground truth''. Ideally, the model of a word should capture the underlying structure of the original data as well as the temporal semantic evolution. The structure will aid neighborhood selection in every timestamp and hence capture how the neighborhood changes. The changes are best represented by performing neighborhood retrieval using the tf-idf model, with the understanding that tf-idf suffers from sparsity- and noise-related issues along with continuity in the time dimension. Thus, in our work we target construction of a model that, for evolution tracking, is good at all of the following aspects: (a) filtering noisy data, (b) keeping the evolution patterns similar to the underlying original data, (c) not as monotonous as when using dynamic embeddings. We again use the previously introduced neighborhood metrics to compare our method with the neighborhoods obtained using tf-idf and dynamic Bernoulli embeddings.

\begin{figure*}
	\centering		
	\includegraphics[width=\textwidth]{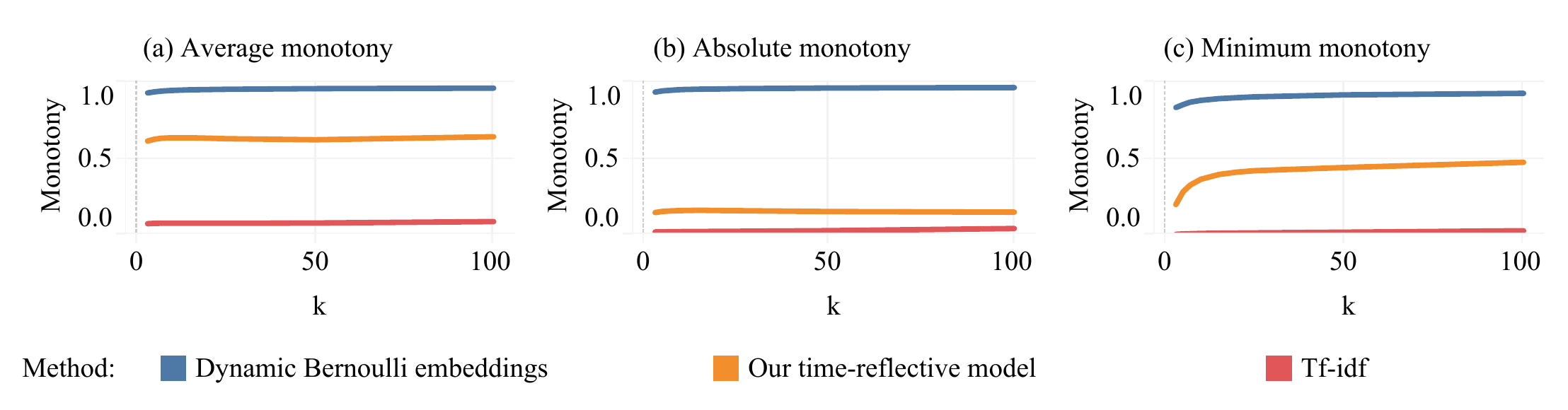}
	\caption{Average(a), absolute(b) and minimum(c) monotony values for different neighborhood sizes ($k$) obtained using three different methods.
		}
	\label{fig:quantitative}
\end{figure*}

Figure~\ref{fig:quantitative} illustrates the trends of the three different methods while changing the size $k$ of the retrieved neighborhoods. As can be seen from the figure, our method achieves results that strike a balance between the highly monotonous results from the dynamic Bernoulli embeddings and the highly unstable results provided by tf-idf. %in contrast to the dynamic Bernoulli embeddings, and that are not very unstable, as the neighborhoods obtained using tf-idf.
The figure also shows the significant differences between the average, absolute, and minimum neighborhood monotony values for different neighborhood sizes when using our method. Of particular interest is the behavior of the minimum monotony which significantly increases as the neighborhood size increases. This is the expected behavior since as the neighborhood size increases, the probability that a high percentage of words change from one timestamp to the other decreases. 

% \section{Parallel version}\label{sec:parallel}
% Optionally we can present the results of running the code in parallel for all the words in the vocabulary, and compare the total run-time with the sum of times required to process every word individually. For this work we used Speedster which has 29 cores. Another option would be to make the code GPU-capable and also compare how much time it takes, but in that case it would have to be sequentially since our GPU can \textbf{probably} only handle the computation for one word at a time.

\section{Conclusions and Future Work} \label{sec:conclusions}

In this work we presented a new time-reflective vector space model representation. This representation allows us to track how the meaning of words changes over time. We compared our model with other text representations, through extensive experiments. Our method obtains a representation that: (1) can track significant changes that are observed within a short period, (2) provides a smooth evolution of the vectors over a continuous temporal vector space, and (3) uses the concept of ``diffusion'' to compensate for the possible sparsity of a dataset. As a future direction, we plan to obtain a lower-dimensional representation of the vectors, as illustrated by \textit{An aspiring approach} of Table \ref{table:comparison}.

\section*{Acknowledgment}
This material is based upon work supported by the U.S. Army Engineering Research and Development Center under Contract No. W9132V-15-C-0006. This work is also supported in part by the National Science Foundation under Grant No. HRD-1242122.


%\bibliographystyle{IEEEtran}
\bstctlcite{BSTcontrol}
\bibliographystyle{IEEEtranS}
%\bibliography{IEEEabrv,Zotero.bib}
\bibliography{IEEEabrv,Config,Zotero}

\end{document}